\documentclass[conference]{IEEEtran}
\IEEEoverridecommandlockouts

\usepackage{cite}
\usepackage{amsmath,amssymb,amsfonts}
\usepackage{algorithmic}
\usepackage{graphicx}
\usepackage{balance}
\usepackage{textcomp}
\usepackage{color,soul}
\usepackage{booktabs}

\usepackage{tikz}
\usepackage{textcomp}
\usepackage{hyperref}

\def\BibTeX{{\rm B\kern-.05em{\sc i\kern-.025em b}\kern-.08em
    T\kern-.1667em\lower.7ex\hbox{E}\kern-.125emX}}

\newcommand\copyrighttext{%
  \footnotesize \textcopyright 2018 IEEE. Personal use of this material is permitted. Permission from IEEE must be obtained for all other uses, in any current or future media, including reprinting/republishing this material for advertising or promotional purposes, creating new collective works, for resale or redistribution to servers or lists, or reuse of any copyrighted component of this work in other works.}

\newcommand\copyrightnotice{%
\begin{tikzpicture}[remember picture,overlay]
\node[anchor=south,yshift=20pt] at (current page.south) {\fbox{\parbox{\dimexpr\textwidth-\fboxsep-\fboxrule\relax}{\copyrighttext}}};
\end{tikzpicture}%
}
\begin{document}

\title{Early Seizure Detection with an Energy-Efficient Convolutional Neural Network on an Implantable Microcontroller}

\author{\IEEEauthorblockN{Maria H{\"u}gle\IEEEauthorrefmark{1}, Simon Heller\IEEEauthorrefmark{2}, Manuel Watter\IEEEauthorrefmark{1}, Manuel Blum\IEEEauthorrefmark{1}, Farrokh Manzouri\IEEEauthorrefmark{3}, Matthias D{\"u}mpelmann\IEEEauthorrefmark{3}}
\IEEEauthorblockN{Andreas Schulze-Bonhage\IEEEauthorrefmark{3}, Peter Woias\IEEEauthorrefmark{2}, Joschka Boedecker\IEEEauthorrefmark{1}}
\IEEEauthorblockA{\IEEEauthorrefmark{1}\textit{Dept. of Computer Science, Faculty of Engineering, University of Freiburg}\\
\{hueglem, watterm, mblum, jboedeck\}@informatik.uni-freiburg.de}
\IEEEauthorblockA{\IEEEauthorrefmark{2}\textit{Dept. of Microsystems Engineering, Faculty of Engineering, University of Freiburg}\\
\{simon.heller, woias\}@imtek.uni-freiburg.de}
\IEEEauthorblockA{\IEEEauthorrefmark{3}\textit{Epilepsy Center, Medical Center, Faculty of Medicine, University of Freiburg}\\
\{farrokh.manzouri, matthias.duempelmann, andreas.schulze-bonhage\}@uniklinik-freiburg.de}
}

\maketitle

\copyrightnotice

\begin{abstract}

Implantable, closed-loop devices for automated early detection and stimulation of epileptic seizures are promising treatment options for patients with severe epilepsy that cannot be treated with traditional means. Most approaches for early seizure detection in the literature are, however, not optimized for implementation on ultra-low power microcontrollers required for long-term implantation. In this paper we present a convolutional neural network for the early detection of seizures from intracranial EEG signals, designed specifically for this purpose. In addition, we investigate approximations to comply with hardware limits while preserving accuracy. We compare our approach to three previously proposed convolutional neural networks and a feature-based SVM classifier with respect to detection accuracy, latency and computational needs. Evaluation is based on a comprehensive database with long-term EEG recordings. The proposed method outperforms the other detectors with a median sensitivity of 0.96, false detection rate of 10.1 per hour and median detection delay of 3.7 seconds, while being the only approach suited to be realized on a low power microcontroller due to its parsimonious use of computational and memory resources. 
\end{abstract}

\begin{IEEEkeywords}
Electroencephalography, Epilepsy, Seizure Detection, Responsive Neurostimulation, Convolutional Neural Network, Low Power Microcontroller
\end{IEEEkeywords}

\section{Introduction}

Epilepsy is one of the most common neurological diseases \cite{Sander1996} and a high percentage of patients with epilepsies are refractory to pharmaceutical therapy \cite{Kwan2000}. A new treatment option for patients with intractable epilepsy is closed-loop brain stimulation \cite{Schulze-Bonhage2016}, with an additional advantage of only short intermittent interventions compared to traditional continuous pharmaceutical therapy. In order to interrupt seizures, a seizure detection algorithm can be used to trigger the intervention using intracranial Electroencephalography (EEG) data. Research on automatic seizure detection started with the objective of reducing workload for the review of long-term recordings in epilepsy monitoring units and moved together with the development of seizure prediction algorithms \cite{Mormann2007} towards the application in implantable devices \cite{Ramgopal2014, Cook2013}. Most of the approaches are based on handcrafted feature selection \cite{Logesparan2012,svm_zheng_2014} or rules designed by experts \cite{Gotman1982}. Driven by the success of deep learning, more recent approaches use deep convolutional networks or recurrent networks  \cite{seizure_prediction_convnet,convnet_acharya, seizure_detection_lstm}.
However, most of these architectures are too demanding for the implementation on an implantable hardware platform.  There are some approaches which are using small convolutional neural networks with only few layers and a small number of weights. The EEGNet of Lawhern et al. \cite{eegnet_2016} is in principle transferable to an implantable hardware. However, the hardware implementation itself was not addressed in their work. Kiral-Kornek et al. \cite{mobile_medical_devices} proposed two architectures for seizure prediction. One convolutional neural network, which was evaluated on a comprehensive data set and another, which can be implemented on a TrueNorth neuromorphic chip from IBM.

In this paper, we propose SeizureNet, which uses efficient layer combinations and has state-of-the-art detection performance. SeizureNet bridges the gap to an implant for seizure detection based on deep learning. To the best of our knowledge, we have designed the first convolutional neural network for seizure detection specifically for an implantable ultra-low power microcontroller. The proposed architecture exhibits low runtime and memory usage, but maintains high sensitivity in combination with a low false positive rate and a short detection delay for a successful stimulation in a later closed-loop application.

After describing the hardware and the dataset in the two following sections, we define the seizure detection problem and explain our detection pipeline, including preprocessing, model architecture, training and performance evaluation in section \ref{sec:methods}. We then show results in section \ref{sec:results} for the actual seizure detection performance, as well as a comparison of hardware properties such as runtime, memory, and energy consumption of our model and four other baselines. We also discuss limitations of the seizure detection device, before concluding in section \ref{sec:conclusion}.

\section{Hardware}
	For the hardware implementation of the network, a low power microcontroller MSP430FR5994 from Texas Instruments is used. Due to its power consumption of 118$\,\mu$A/MHz in active mode and 0.5$\,\mu$A in standby mode, it is suitable for the application in an implantable device where a heating of the surrounding tissue must be avoided. 
A further great advantage of the MSP430FR series is its ferromagnetic nonvolatile memory (FRAM). With a low-power consumption and fast write speed, a swift storage of hidden layer activations of a neural network can be implemented. However, the FRAM also limits the maximum clock speed of the controller as its reading speed is limited to $8$\,MHz. It is possible to run the controller with higher clock speeds but only with additional wait states for the CPU leading to a lower power efficiency.
Another useful feature for the implementation of convolution layers is the $32$-bit hardware multiplier of the controller, enabling power efficient multiply and accumulate (MAC) operations without CPU intervention.

\section{Dataset}
The  dataset used is the \emph{Epilepsiae} database \cite{Klatt2012}, containing long-term continuous intracranial EEG data. We evaluate our approach on 24 patients. Each recording has a duration between five and eleven days and contains the measurement of approximately 100 intracranial and scalp electrodes originally sampled with or resampled to $f_s = 256\,$Hz. During the two weeks, the evaluated patients had between $6-92$ seizures. To limit the amount of data for our experiments, we consider 100 minutes segments of the recordings around the seizures. 

For every patient, we consider a subset of $E=4$ electrodes, which are selected a priori by expert epileptologists to cover the seizure onset zone(s). In case that less than four electrodes display the initial ictal EEG pattern, neighboring channels are included for seizure detection. The total number of electrodes is limited due to hardware limitations. 

\section{Methods}
\label{sec:methods}
\subsection{Seizure Detection Problem}
Seizure detection can be modeled as time-series classification, where we classify ictal phases (seizures) and interictal phases (non-seizures).
To create the inputs for the convolutional network, we process sliding windows over the EEG data $D \in \mathbb{R}^{T \cdot E}$, where $T$ is the recording duration and $E$ is the number of electrodes. $1$ second samples are created as input features $x_t\in R^I$ with input dimensionality $I=\lfloor T/S \rfloor  \cdot E \cdot f_s$ at time point $t$, where $f_s$ is the sampling frequency and $S$ the stride.
The corresponding labels are $y \in \{0,1 \}^{\lfloor T/S \rfloor}$, where $y_i = 1$ indicates a seizure at the end of sample $i$ and $y_i = 0$ an interictal sample.  The window lengths are chosen to keep the runtime of a forward pass low. To train our models, an overlap of $50\%$ is used, which equals a stride of $S=0.5$ seconds. Due to the hardware runtime limitations, we evaluate our models with a stride of 1$\,$s.

\subsection{Preprocessing}
Compared to the conventional scalp EEG, intracranial EEG is less prone to artifacts like the pick-up of the electrocardiogram or electromyogram. However, a careful removal of noise and drifts in the data can facilitate the subsequent pattern classification task. Multiple preprocessing steps are performed to remove signal components which are not carrying relevant information and to adapt the signal statistics to intra-individual fluctuations. The power line noise at $50\,$Hz is removed by applying a notch-filter on the raw EEG data to exclude this frequency. Subsequently, a highpass filter ($\ge 0.1\,$Hz) removes slow drifts. The data is then rescaled by dividing through the rolling 10 minutes standard deviation to account for non-stationarity in the source. Preliminary experiments showed this setting to perform well. The rolling mean and standard deviation can be computed for a time point $t$ and window size $N$ as:
\begin{align*}
\mu_t^N(x_{t-N:t}) & = \frac{1}{N} \sum_{i =t-N}^{t} x_i \\
\sigma_t^N(x_{t-N:t}) & = \frac{1}{N}  \sum_{i =t-N}^{t}  \left(x_i - \mu_t^N(x_{t-N:t}) \right)^2
\end{align*}

\noindent The input data is then normalized as follows:
$$ \displaystyle \hat x_t = \tanh\left(0.2 \cdot \frac{x_t}{\sigma^{N_w}_t(x_{t-N_w:t})}\right)$$
where $N_w = 153600$, which equals 10 minutes of data. 
We normalize the scaled and standardized data via the hyperbolic tangent to reduce the influence of outliers and artifacts while maintaining a quasi-linear relation for most of the input distribution.

\subsection{Model Architecture}

In order to find a good model architecture, we evaluated the runtime and memory requirements for various layer types like convolutions, dense layers, pooling layers and activations. The architecture of SeizureNet is shown in Table \ref{lab:tab:seizurenet}. The proposed network is a deep convolutional network with alternating convolutional and pooling layers.

\begin{table}[b!]
\caption{Architecture of SeizureNet for $E$ input electrodes.}
\center
\label{lab:tab:seizurenet}
\begin{tabular}{ c c c c  }
  \toprule
  Layer & Operation & Output \\
  \midrule 
     &  Input  $(E \times 256)$  & $E \times 256 \times 1$  &\\
  1  & $20 \times \text{Conv2D } (E \times 17)$ & $1 \times 240 \times 20$ &  \\
  2   & $\text{MaxPool2D } (1 \times 4)$ & $1 \times 60 \times 20$ & \\
    & Dropout (0.2) & $1 \times 60 \times 20$  &\\
    3 & $10 \times \text{Conv2D } (1 \times 5)$ & $1 \times 56 \times 10$  & \\
  4   & $ \text{MaxPool2D } (1 \times 4)$ & $1 \times 14 \times 10$ & \\   
    & Dropout (0.2) &  $1 \times 14 \times 10$  &\\
     5 & $10 \times \text{Conv2D } (1 \times 5)$ & $1 \times 10 \times 10$  & \\
   6 & $ \text{MaxPool2D } (1 \times 2)$ & $1 \times 5 \times 10$ & \\   
    & Dropout (0.2) &   $1 \times 5 \times 10$   &\\
         7 & $10 \times \text{Conv2D } (1 \times 5)$ & $1 \times 1 \times 10$  & \\  
    & Dropout (0.2) &   $1 \times 1 \times 10$   &\\
       8 & $1 \times \text{Conv2D } (1 \times 1)$ & $1 \times 1 \times 
1$  & \\ 
  9     & Sigmoid & $1$  & \\  
  \midrule
 \multicolumn{3}{c}{Total Number of Parameters:     3,621 }\\
\bottomrule
\end{tabular}
\end{table}

Lawhern et al. \cite{eegnet_2016} proposed convolutions over electrodes in the first layer. They used kernels of size $(E, 1)$, which is similar to approaches such as common spatial patterns \cite{blankertz_csp_2008}. We extend this by convolving over the electrodes and time, so that we can learn spatio-temporal patterns efficiently in one layer.
In the last layer, we use $1 \times 1$ convolutions instead of a fully-connected layer. This was introduced in \cite{networkinnetwork_lin_2013} and is a parameter-efficient way to reduce dimensions \cite{springenberg_allnet_2015}.
Rectified Linear Units (ReLu) are used in all hidden layers. Further, batch normalization is used after the convolutions. During training, dropout regularization is applied to reduce overfitting.

\subsection{Training}
To deal with the high imbalance of ictal and interictal samples, we use an oversampling technique. Mini-batches are created by randomly picking ictal samples with probability $p$, and interictal samples otherwise. 
In order to learn to detect seizures as early as possible, we weight the ictal samples in the loss function according to their distances to the seizure onset.
The weights decrease linearly from $1.0$ for the onset to $0.0$ for the seizure offset.

In all experiments, the patient-specific models are trained. For evaluation, we use 3-fold cross validation. Each model is trained with a batch size of $256$ for $1.5 \cdot 10^5$ steps (batches of samples), with a sampling probability $p=0.1$ for the number of seizures in each batch. For optimization, we use Adam \cite{DBLP:journals/corr/KingmaB14} with a learning rate of $10^{-3}$ and the binary cross-entropy loss.

\subsection{Detection Performance Evaluation}
\label{lab:sec:perfromance_evaluation}
It is non-trivial to evaluate a seizure detection system. Mainly, three objectives should be optimized: 

\begin{itemize}
\item The \emph{sensitivity} is defined as the ratio of actually detected seizures to the total number of seizures.
\item The \emph{detection delay} is calculated as the mean delay over all detected seizures. For each detected seizure, the delay is defined as the expired time between the electrographic seizure onset identified through visual inspection by a domain expert, and the first algorithm-based detection of the seizure.
\item The \emph{false positive rate} is the number of false detections per hour (\emph{fp/h}). 
\end{itemize}

\subsection{Approximations}

\label{sec:approximations}
For our hardware implementation, we have to approximate the rolling 10 minutes standard deviation due to memory limitations. We first approximate the rolling 10 minutes mean $\mu^{N_w}_t$ by computing a grand mean $\tilde \mu_t^{N_w}$ over $1$-second means $\bar \mu_k$. Doing so, only $K = 600$ means have to be stored for the targeted duration, while introducing a negligible error by ignoring the most recent samples until they form a new 1 second mean $\bar \mu_1$. The approximated mean can then be used to compute the 10 minutes standard deviation $\tilde \sigma^{N_w}_t$:
\begin{align*}
\bar \mu_k & = \mu_{t - k \cdot f_s}^{f_s} \left( x_{t - k \cdot f_s : t - (k - 1) f_s} \right) \\
 \tilde \mu_t^{N_w} & = \mu_t^K \left( \bar \mu_{1:K} \right) \\
\left(\tilde \sigma_t^{N_w}\right)^2 & = \frac{1}{N_w} \sum_{i =t-N}^{t} \left(x_i - \tilde \mu^{N_w}_t\right)^2 \\
\end{align*} with $k \in \{1, \ldots, K\}$.

Further, we use a linear approximation of the hyperbolic tangent in the preprocessing to avoid the need for a lookup-table: 
$$\text{lintanh}(x) :=  \begin{cases}
x /1.2 & -1.2 \le x \le 1.2  \\
1 & \, x > 1.2 \\
-1 & \, x < -1.2
\end{cases}$$

\subsection{Comparison to other Approaches}
For seizure detection, it is hard to compare approaches without evaluating on the same framework, due to factors like different evaluation methods, different datasets, omitted patients and patient-specific or across-patients training. Hence, we reimplemented all our baselines. We compare the performance of our architecture to three other convolutional neural networks, which we find most similar to our approach. In order to evaluate fairly, we use the same preprocessing and the sampling method for all convolutional networks. The architectures of the network baselines are shown in Table \ref{lab:tab:baselines}. Further, we compare to the performance of a support vector machine approach from the literature using handcrafted features \cite{svm_zheng_2014}.

\begin{table}[b]
\caption{Architecture of the baseline networks.}
\label{lab:tab:baselines}
\begin{tabular}{c |c |c }
 \toprule
 EEGNet & Kiral-Kornek et al. & Acharya et al.\\
\midrule
 \multicolumn{1}{c |}{
 %row 1
 Input  $(E \times 256)$  } 
 &Input  $( 32 \times 32 \times E )$  
 & Input  $( E \times 4097 )$   \\
 %row 2
  $16 \times \text{Conv2D } (E \times 1)$ & 
  $16 \times \text{Conv2D } (3 \times 3)$  & 
  $4 \times \text{Conv2D } (E \times 6)$ \\
  %row 3 
  Transpose &  
  $ \text{MaxPool2D } (2 \times 2)$ &
  $ \text{MaxPool2D } (1 \times 2)$  \\
  %row 4 
 Dropout $(0.25)$ & 
 Dropout $(0.7)$&
  $4 \times \text{Conv2D } (1 \times 5)$ \\
   %row 5 
   $4 \times \text{Conv2D } (2 \times 32)$  & 
  $32 \times \text{Conv2D } (3 \times 3)$&
  $ \text{MaxPool2D } (1 \times 2)$  
  \\
  %row 6
  $\text{MaxPool2D } (2 \times 4)$  &
  $ \text{MaxPool2D } (2 \times 2)$ &
   $10 \times \text{Conv2D } (1 \times 4)$ 
  \\
   %row 7 
 Dropout $(0.25)$   & 
  Dropout $(0.7)$&
  $ \text{MaxPool2D } (1 \times 2)$    
  \\
  %row 8 
    $4 \times \text{Conv2D } (8 \times 4)$ & 
    $32 \times \text{Conv2D } (3 \times 3)$& 
   $10 \times \text{Conv2D } (1 \times 4)$ 
    \\
  %row 9 
  $ \text{MaxPool2D } (2 \times 4)$  &  
  $ \text{MaxPool2D } (2 \times 2)$ &
  $ \text{MaxPool2D } (1 \times 2)$    
  \\
  %row 10 
  Dropout $(0.25)$ & 
  Dropout $(0.7)$ & 
   $15 \times \text{Conv2D } (1 \times 4)$  \\
  %row 11 
 $1 \times \text{Conv2D } (1 \times 1)$ & 
 Dense  $(32)$& 
  $ \text{MaxPool2D } (1 \times 2)$  \\
 %row 12
 GlobalMaxPool2D  &
 $ \text{Dropout} (0.5) $ & 
  Dense $(50)$\\
   %row 13
  & &  Dense  $(20)$\\
   %row 14
   \midrule
 \multicolumn{3}{c}{Sigmoid Output Layer}\\ 
 \midrule
 \multicolumn{3}{c}{Total Number of Parameters:}\\
  \multicolumn{1}{c}{957} & \multicolumn{1}{c}{15,665} & \multicolumn{1}{c}{96,220}\\
 \bottomrule 
\end{tabular}
\end{table}

\subsubsection{EEGNet}
The first baseline is EEGNet, a small convolutional network by Lawhern et al. \cite{eegnet_2016}. The EEGNet shows robust performance across four different brain-computer interface classification tasks. It consists of three convolutional layers and two max pooling layers, where the first convolution estimates a set of spatial filters over the electrodes. To adapt their approach to our framework, we replace the softmax regression output layer by a sigmoid activation.

\subsubsection{Kiral-Kornek et al.}
In \cite{mobile_medical_devices}, a convolutional network is evaluated for patient-specific seizure prediction based on spectrograms. To compare with their architecture, we use their approach for seizure detection instead of prediction. They proposed a network consisting of three alternating convolutional layers and three max-pooling layers using $ 32 \times 32$ spectrograms as input. As activation function, they use an Exponential Linear Unit (ELU) \cite{elus}.
To provide a proof-of-concept that they can implement their seizure prediction system on a low-power system, they adapt their network architecture to a 18-layer binary neural network, consisting only of convolutional layers and dropout. This architecture can run on the IBM \emph{TrueNorth Neurosynaptic System} chip \cite{truenorth_chip_esser_2016}. However, because their network uses binary weights, they need more layers and thus more parameters to reach the same precision. In total, they use over 4.2 million parameters, which would cost more than $525\,$kB for binary weights. Because this already exceeds our memory limit, we only use their small architecture as a baseline. To adapt this approach to seizure detection, we use 1 second windows instead of 30 second windows and Short-Time Fourier Transformation (STFT) to generate the spectrograms. Further, we focus on the comparison of the network architectures and thus don't include time of day as an input feature to keep the input consistent across methods. 

\subsubsection{Acharya et al.} In \cite{convnet_acharya}, a 13-layer convolutional neural network is  trained across patients on one electrode to classify interictal, ictal, and preictal phases (phases directly before a seizure).  They trained their model on a subset of the \emph{Epilepsiae} database, consisting of 5 epileptic patients and additionally 5 healthy subjects. The architecture takes an input window of length $4097$, which equals to $16\,$s for a sampling rate of $256\,$Hz. As activation function, they use a Leaky ReLu. To stick to our framework, we adapt their network so that it uses $4$ electrodes as input by changing the first convolutional layer from a $6 \times 1$ convolution to $6 \times 4$. Further, we use a sigmoid output layer. For training, an overlap of $50\%$ is used. As for the other approaches, the performance is evaluated with a $1$ second stride.

\subsubsection{SVM}
Besides deep learning approaches, we compare our method to the support vector machine (SVM) proposed in \cite{svm_zheng_2014}. For feature extraction, they use the EMD algorithm \cite{huang_emd_algorithm}, which is a signal processing method for nonlinear and non-stationary time series. The SVM uses the variance of Intrinsic Mode Functions (IMFs) as input features and a Radial Basis Function (RBF) as kernel. A post processing step is applied, which classifies a seizure only if the SVM has classified a number of consecutive epochs as epileptic activity. In the original study, eight consecutive epochs were required for detecting a seizure. Since this would lead to a high detection delay for online detection steps of one second, we skip this step. Further, they use only 20 seconds of ictal and 100 seconds of interictal data for training. In contrast to that, we use all available seizures of the training set and downsample the remaining interictal data to the ratio proposed in the original paper.  In our experiments, we use the first three extracted IMFs as features for the SVM.  

\section{Results}
\label{sec:results}
In order to evaluate our approach, we first compare the detection performance for all architectures.  Further, the ROC AUC score is computed to show the discriminative properties of the models. Since the AUC score is independent of the classifier threshold, it is a suitable additional metric to compare the different architectures. However, since seizures only have to be detected once but as fast as possible, the per-sample sensitivity and thus the AUC score of the classifier is not as important as the other metrics.
After that, we show how the performance metrics can be influenced by varying the classification threshold. This offers the possibility of manual (re-)adjustment of the detection device during a study with real patients. Then, we investigate the effect of the preprocessing approximations.
Finally, we evaluate all network-based models regarding their memory and speed for our hardware implementation. For this purpose, we analyze the requirements for the preprocessing and the successive layers of the networks.  

\begin{figure}[h]
\center
  \includegraphics[width=0.45\textwidth]{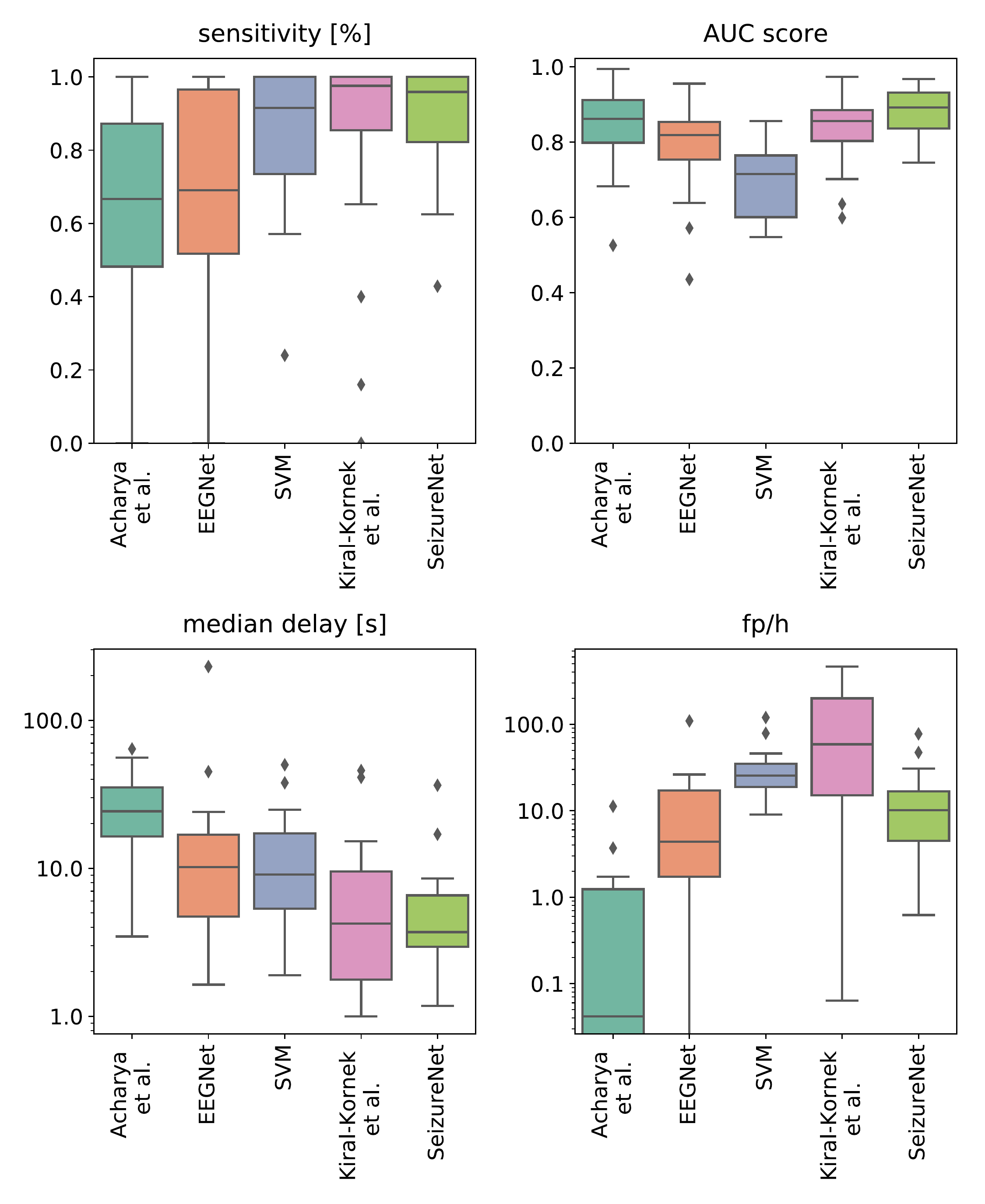}
\caption{Detection performances over all patients for a classifier threshold of $0.5$. Median delay and false positive rate are shown in logarithmic scale.}
\label{fig:performance_all}
\end{figure}

The detection performance of our model and all baselines is shown in Fig. \ref{fig:performance_all} for a classifier threshold of $0.5$. To show the overall performance, we calculate the metrics separately for each patient and summarize the distribution in the figure. SeizureNet has robust and high sensitivities for all patients. With a median sensitivity of $0.96$, it outperforms all other architectures except for the network of Kiral-Kornek et al., which has a median sensitivity of $0.98$. Their network, however, shows extremely high false positives rates, with outliers up to $464$ false positives per hour. Because of the large window size of $16\,$s, the lowest median $fp/h$ of $0.04$ is achieved by the network of Acharya et al. However, this window leads to a high median delay of $24.3\,$s (unacceptable for \textit{early} seizure detection applications) and a low sensitivity of $0.67$. The second best false positive rate is achieved by EEGNet due to its low sensitivity. The SVM has a good median sensitivity, but a higher delay and a higher false positive rate than SeizureNet. 
With the best median AUC score of $0.89$ and a good balance between false positives and delay, SeizureNet shows the best and most robust detection performance.

\begin{figure}[h]
\center  \includegraphics[width=0.5\textwidth]{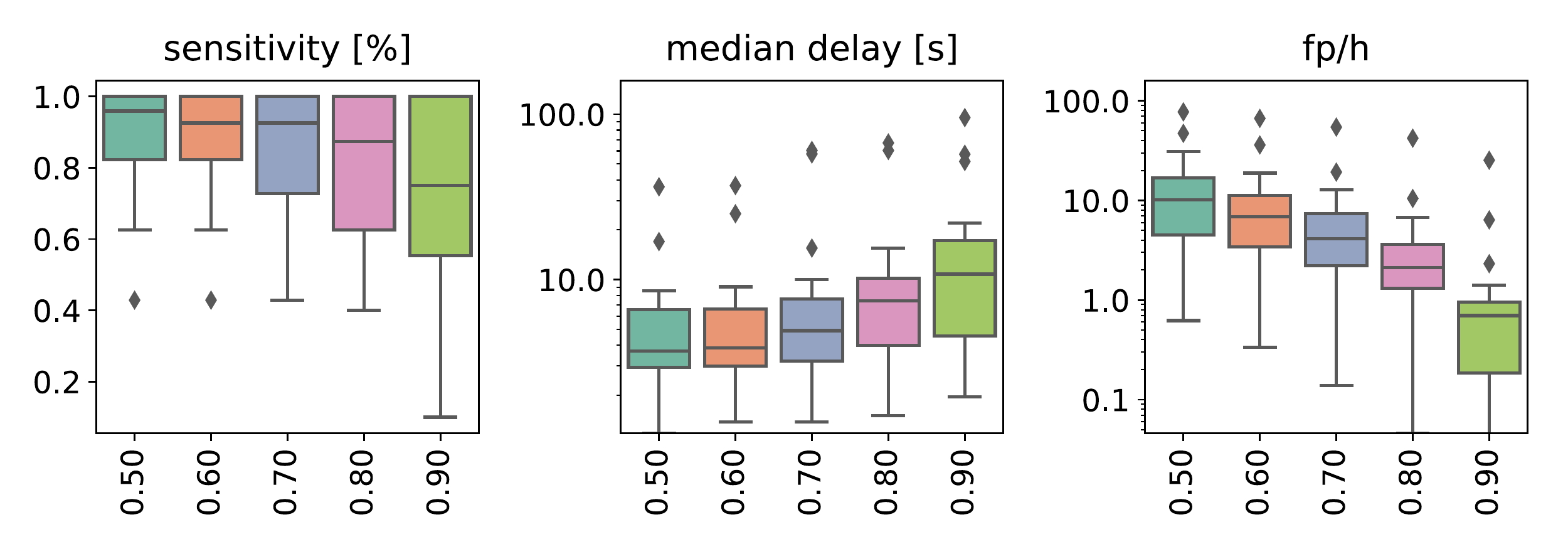}
\vspace*{-4mm}
\caption{Influence of classifier threshold on the detection performance of SeizureNet. Median delay and false positive rate are shown in logarithmic scale.}
\label{fig:classifier_threshold}
\end{figure}

Fig. \ref{fig:classifier_threshold} shows the influence of the classifier threshold on the performance metrics. 
Higher thresholds correlate with reduced sensitivity and increased delays, but also prove to be less prone to false alarms.
The highest median sensitivity of $0.96$ is achieved with a classifier threshold $0.5$, with a respective median delay of $3.7\,$s and $10.1$ false positives per hour. A low median $fp/h$ of $0.7$ can be achieved with a high classifier threshold of $0.9$. However, this threshold decreases the sensitivity to $0.75$ and increases the median delay to $10.8\,$s.
Of course, the classifier threshold can also be varied for the other architectures (for the SVM, the proposed postprocessing step can be used instead). However, as indicated by the high AUC score, SeizureNet has the best performance independent of the threshold.

\subsection{Approximations}

\begin{figure}[t]
\center
  \includegraphics[width=0.5\textwidth]{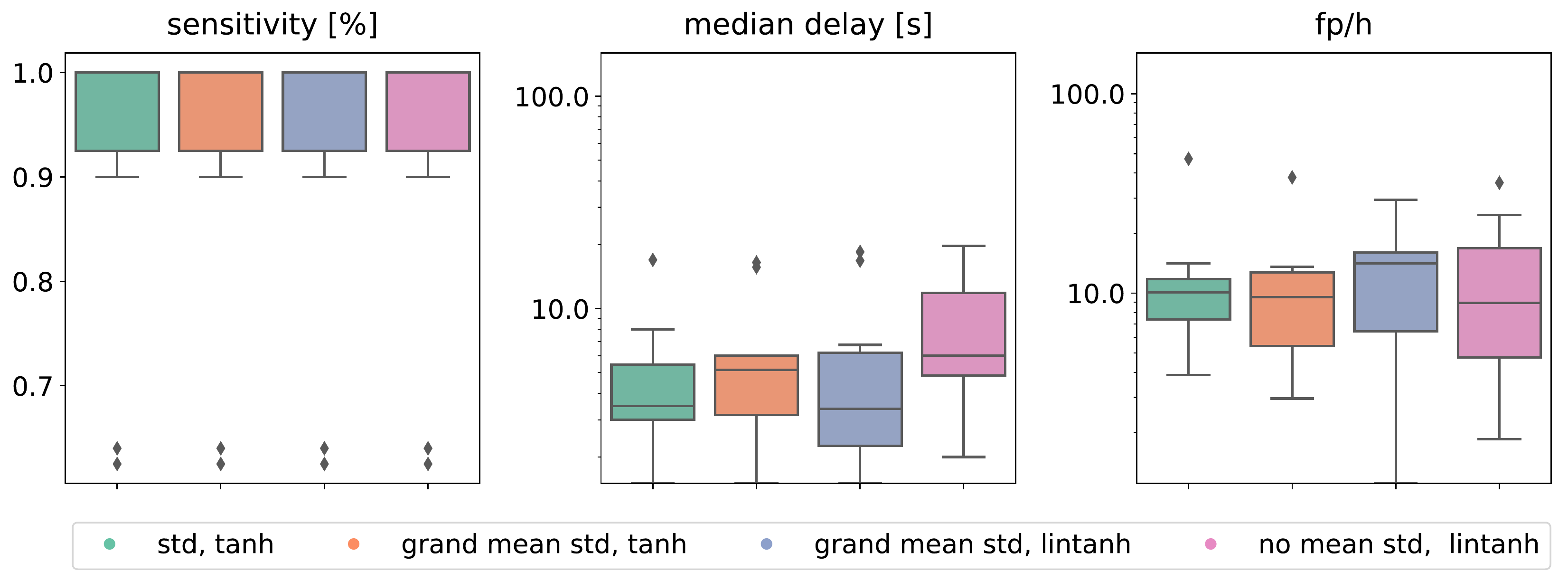}
\caption{Detection performance of the proposed preprocessing and various approximations evaluated on a subset of 10 patients.}
\label{fig:approximations}
\end{figure}

Fig. \ref{fig:approximations} illustrates the performance loss due to preprocessing for the proposed approximations for 10 patients. We compare to another approximation, where we replace the rolling mean with 0. The used highpass filter roughly centers the signal around the zero line which allows us to dispense with a mean estimation and the standard deviation can be approximated by $$\left(\hat \sigma_t^{N_w}\right)^2 \approx \frac{1}{N_w} \sum_{i =t-N_w}^{t}x_i^2 $$

In our experiments, using the 10 minutes standard deviation and the hyperbolic tangent shows the best results. The best approximation is using the grand mean with a median delay of $3.4\,s$ and a $fp/h$ of  $14.1$ and an AUC score of $0.94$. The zero mean approximation offers a big advantage in runtime and memory reduction, however, it increases the median delay considerably to $6\,s$.

\subsection{Hardware Requirement Analysis}

The theoretical runtime and memory requirements of all convolutional networks for our hardware implementation are shown in Fig. 4.  Besides SeizureNet, only the network of Kiral Kornek et al. would actually be implementable on our microcontroller. However, the preprocessing of this network is extensive due to the STFT  and has a 38\% higher runtime. The high runtimes of EEGNet and Acharya et al. make an implementation on our realtime detection device impossible. For EEGNet, this is caused by zero-padded convolutions, which are not reducing the dimensionality of the input. The runtime for the network of Acharya et al. is mainly affected by the large input window.

The required memory of the hardware implementation is specified by the number of parameters of the networks, a buffer for the rolling window of the preprocessing, lookup-tables and two buffers, saving the inputs and outputs of the hidden layers. The size of the buffers is determined by the layer with the largest input and output size. Regarding the memory, all networks besides the network proposed of Acharya et al. are theoretically implementable on our device. The limiting factor of this network is the use of fully-connected layers, which require 74\% of the total memory.

Besides SeizureNet, none of the networks were actually implemented on hardware platforms, so we cannot compare  hardware efficiency and power consumption. Only Kiral-Kornek et al. implemented an adapted network on the IBM \emph{TrueNorth} chip. This chip has a consumption of $40-70\,$mW to power 1 Mio neurons, resulting in a power consumption of $40 - 70\,$ nW per neuron. With $107400$ neurons, a power consumption of $4.3 - 7.5\,$mW can be estimated for the adapted network. For the preprocessing and forward pass of SeizureNet, we measured a power consumption of $850\,\mu W$ which is 5 to 8.8 times lower.

\begin{figure}[t]
\center
  \includegraphics[width=0.47\textwidth]{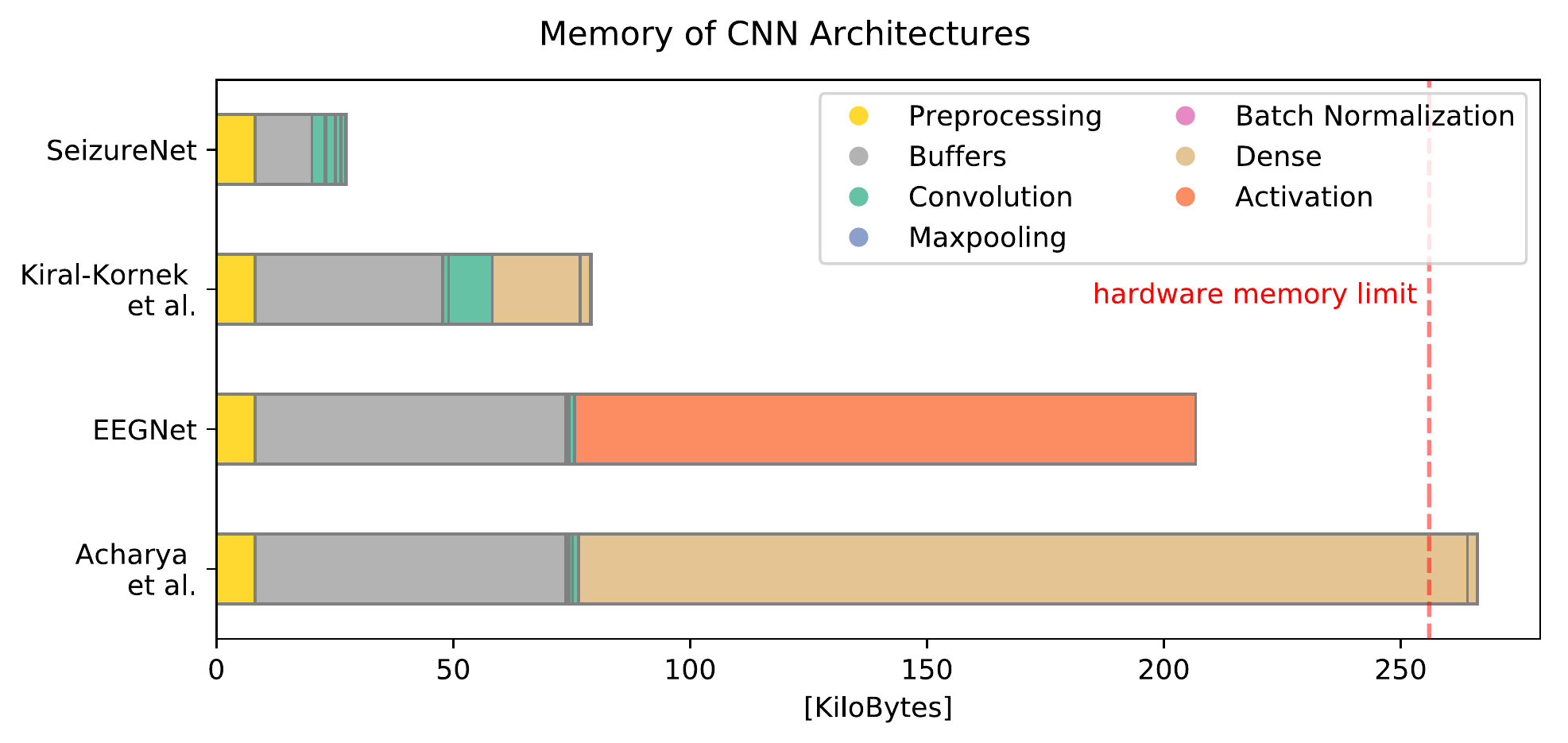}
  \vspace{1cm}
 \hspace*{0.2mm} \includegraphics[width=0.47\textwidth]{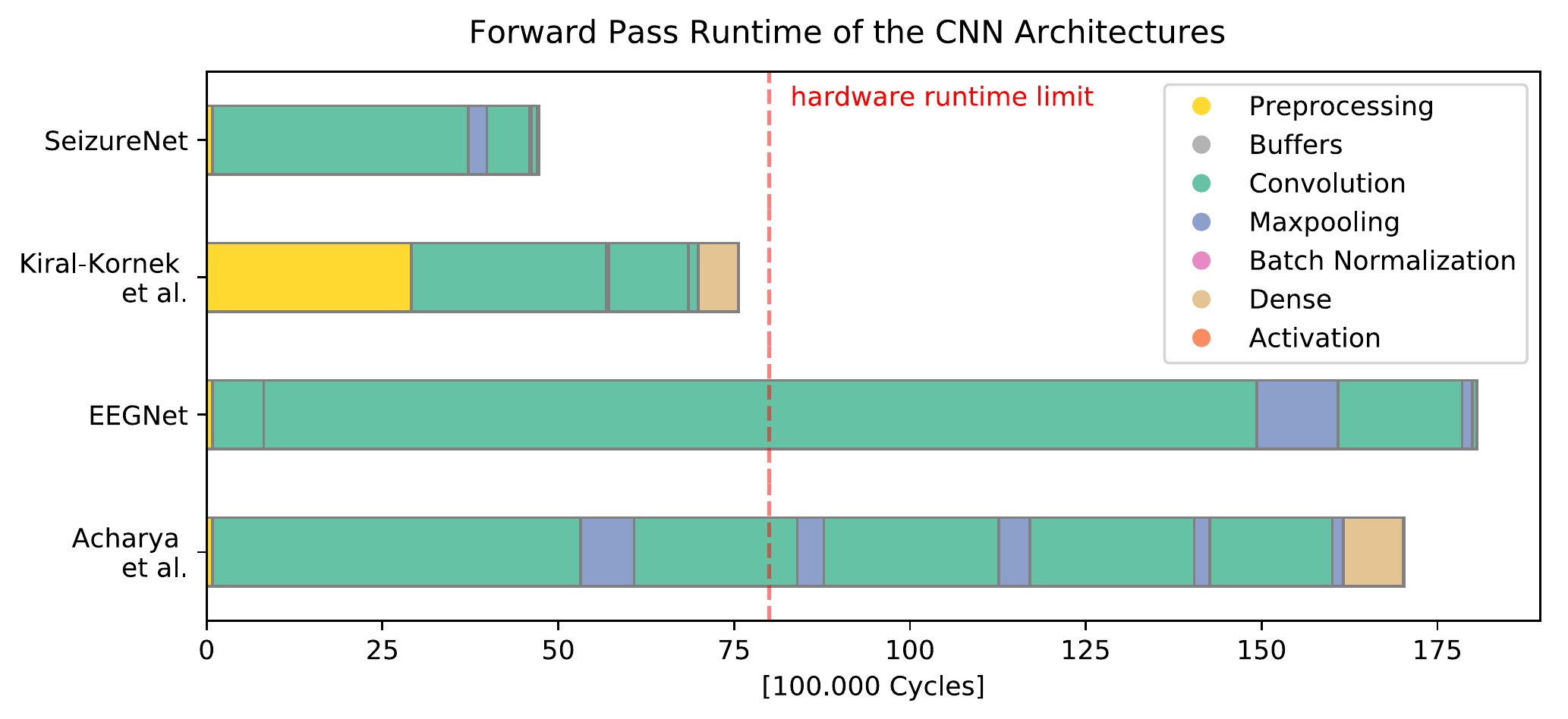}
  \vspace*{-1.0cm}
\caption{Memory (top) and runtime (bottom) requirements for SeizureNet and the baseline architectures. Runtime blocks are ordered according to their execution in the forward pass. Layers with few parameters or cycles are not visible.}
\label{fig:memory}
\end{figure}

The SVM approach was excluded from this comparison as non-parametric approaches behave differently for each patient. While test-time predictions scale linearly in time and memory with the number of support vectors, the required amount grows rapidly with the number of seizures considered in the training data. Due to the varying computational demands for each patient, it is difficult to find a fair setting for a comparison against the network approaches that have a constant load across patients. Additionally, the preprocessing in \cite{svm_zheng_2014} would require simplification to be efficiently implementable and change the method substantially.

\subsection{Limitations of the Approach}

\begin{figure}[t]
\center
\includegraphics[scale=0.23]{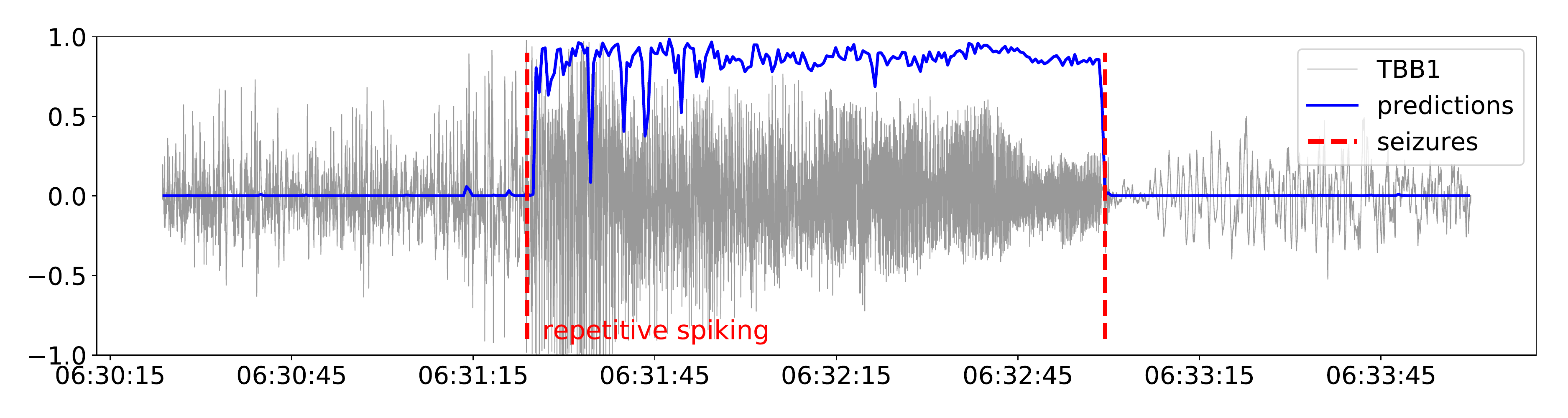}
\caption{Early detected seizure with the repetitive spiking onset pattern: predictions one minute around the seizures and the normalized electrode signal of one electrode.}
\label{fig:early_detection_samples}
\end{figure}

Tuning the networks to early and sensitive detection of electroencephalographic seizure patterns (Fig. \ref{fig:early_detection_samples}) occurs at the cost of higher false positive rates. Several components can have contributed to this: a) selection of short EEG segments of $1\,$s decreases detection delay but increases chances to falsely classify artefacts as ictal patterns; longer windows of analyses can ameliorate this (note the low FP rate in \cite{convnet_acharya} using a $16\,$s window); b) subclinical ictal electroencephalographic patterns resembling ictal patterns by definition, but not accompanied by clinical phenomena can be detected which may occur more frequently than clinically manifest seizures \cite{Feldwisch-Drentrup2011}. These detections should not be called false detections, and there may even be reasons to use such detections to trigger interventions in a closed-loop device setting. Integration of more EEG channels may allow for an estimation of the probability of clinical correlates of the ictal event.

\begin{figure}[t]
\center
  \includegraphics[scale=0.23]{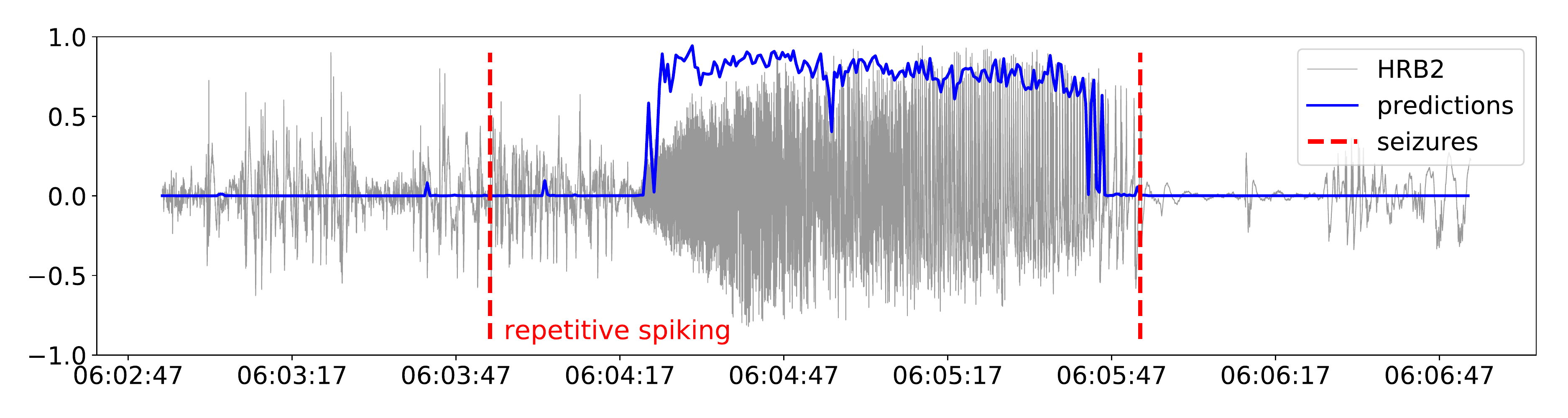}
  \includegraphics[scale=0.23]{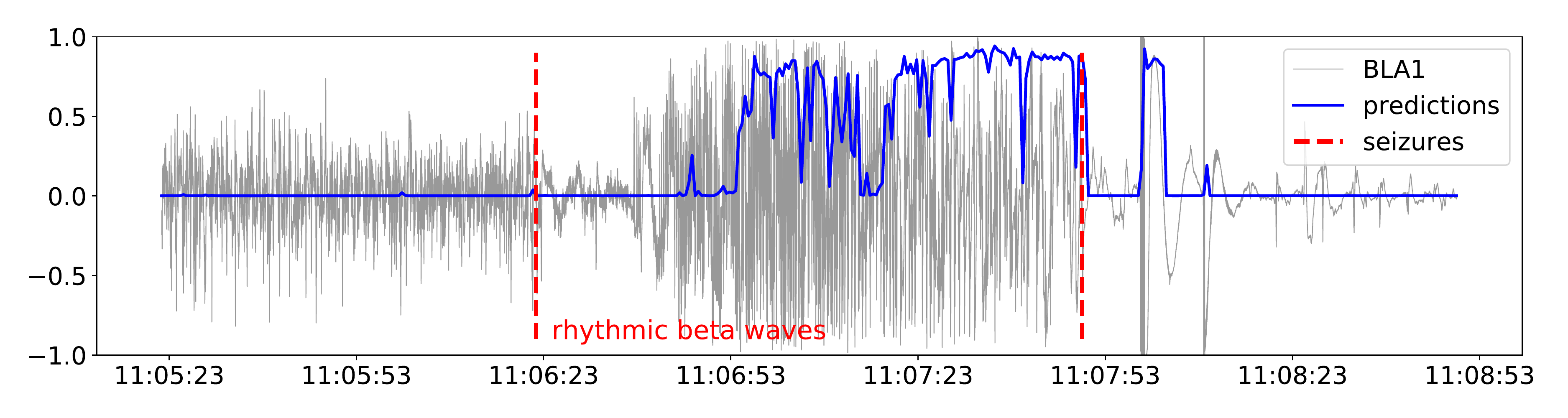}
  \includegraphics[scale=0.23]{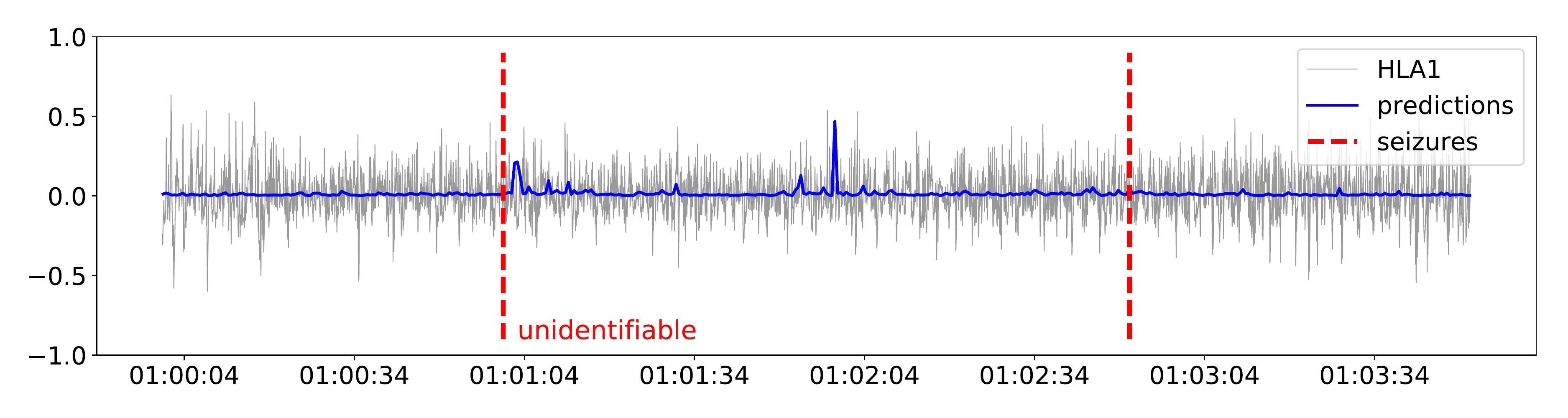}
  \caption{Late detections (top, middle) and missed seizure (bottom): predictions one minute around the seizures and the normalized electrode signal of one electrode.}
\label{fig:late_detection_samples}
\end{figure}

Expert review of some missed seizure showed epileptic auras with unclear electrographic correlates, which allowed neither clear visual nor algorithmic detection (Fig. \ref{fig:late_detection_samples} bottom). Additional investigations are needed to identify the patterns relevant for seizure detection during the evolution of the electrographic ictal event, which not always coincide with the seizure onset pattern, but rather with alterations of the ictal pattern in the course of recruitment and spread (Fig. \ref{fig:late_detection_samples} top and middle). Remarkably, SeizureNet had also good detection latencies for seizures with onset patterns,  which are difficult to detect, for example for amplitude depression \cite{Meier2008}.

\section{Conclusion}
\label{sec:conclusion}
We have introduced SeizureNet, a convolutional neural network for online seizure detection with state-of-the-art performance. Empirical evaluation of our approach demonstrates its suitability for practical realization on an implantable low-power microcontroller for clinical applications. The considered approximations to preprocessing and architecture choices preserve performance sufficiently while leaving over computational resources for further improvement. Candidates for future work in this direction are the distillation of network ensembles \cite{knowledge_distillation}  or the improvements in quantized, low-precision neural networks \cite{quantization}.

\section*{Acknowledgment}
This work was supported by BrainLinks-BrainTools Cluster of Excellence funded by the German Research Foundation (DFG, grant number EXC 1086).

\balance
\bibliography{literature}{}
\bibliographystyle{unsrt}

\end{document}